# Quality4.0[1] - Transparent product quality supervision in the age of Industry 4.0

Jens Brandenburger[1], Christoph Schirm[2], Josef Melcher[2], Edgar Hancke[2], Marco Vannucci[3], Valentina Colla[3], Silvia Cateni[3], Rami Sellami[4], Sébastien Dupont[4], Annick Majchrowski[4] and Asier Arteaga[5]

[1] VDEH-Betriebsforschungsinstitut, Sohnstr. 65, 40237 Düsseldorf, Germany
[2] thyssenkrupp Rasselstein GmbH, Koblenzer Str. 141, 56626 Andernach, Germany
[3] Scuola Superiore di Studi Universitari e di Perfezionamento Sant'Anna,
Piazza Martiri della Liberta 33, 56127 Pisa, Italy
[4] CETIC - Centre d'Excellence en Technologies de l'Information et de la Communication,
Avenue Jean Mermoz 28, 6041 Charleroi, Belgium
[5] Sidenor I+D, Barrio Urgate, 48970 Basauri, Spain

**Abstract.** Progressive digitalization is changing the game of many industrial sectors. Focusing on product quality the main profitability driver of this so-called Industry 4.0 will be the horizontal integration of information over the complete supply chain. Therefore, the European RFCS project "Quality4.0" aims in developing an adaptive platform, which releases decisions on product quality and provides tailored information of high reliability that can be individually exchanged with customers. In this context Machine Learning will be used to detect outliers in the quality data.
This paper discusses the intermediate project results and the concepts developed so far for this horizontal integration of quality information.

## 1 Introduction

In a world where steel products can be acquired through platforms like Alibaba.com and the steel market is flooded with cheap steel from Chinese overcapacity, European Steel producers urgently need differentiation as distance or referencing are not a protection anymore [1].

From steel customers' point of view, one main reason for the decision for a specific supplier is trust in the fact that the delivered product fulfills his individual requirements. Consequently, only if the European Steel Industry succeeds to win customer-trust and solidifies client intimacy, it will achieve a durable competitive advantage and thus reduce pressure from world-wide imports. This need especially holds for markets like automotive industry were product and data traceability are gaining im-portance. But how to gain customer-trust and intimacy? Keys to success in this context are communication and information exchange, as the only feasible way to make steel customers rely on assured product quality is to share related quality information. On the other hand, customer-feedback can help the supplier as well to improve his own processes, so a bidirectional exchange of quality information would be a major advantage for both sides. In the context of recent discussions about Industry 4.0 this approach is referred to as the "horizontal integration over complete supply chains" and identifies one of the four main aspects characterizing this vision of digital transformation of manufacturing processes [2].

[1] This project has received funding from the Research Fund for Coal and Steel of the European Union (EU) under grant agreement No 788552.

## 2    Problem description

The digitalisation process often referred to as "Industry 4.0" has not fully reached the industrial practise in European Steel Industry yet. Although recently a great development of general-purpose IT-platforms and solutions for industrial data handling and treatment could be observed, in a recent study 62.6% of the steel customers assessed the digitalization level of the Steel Industry still as low [3].

This is because general-purpose IT-Platforms, such as, for instance, MindSphere from Siemens[2], Predix from General Electric[3] and IBM Watson[4] or Open Source platforms like OpenIoT, Eclipse Kura or Kaa usually forget that a lot of work and process knowledge is mandatory to adapt existing products to complex industrial value-chains as formed by the steel manufacturing. As stated in [1], only if process knowledge is combined with IT expertise, value creation by means of digitalisation becomes possible. When thinking about the horizontal integration of quality information over the complete supply chain, it becomes obvious very quickly, that this idea implies much more than just the definition of some standard IT-interfaces and data formats for the exchange of quality data. At first, data characterizing the quality of a product is by far the most critical information gathered during the steel production process, especially with respect to the supplier-customer relationship. To exchange just this crucial kind of information, immediately raises the demand on information reliability to the utmost possible level. To share wrong quality information may cause severe customer uncertainty and long-term damaged customer confidence and should be avoided at all costs.

Consequently, a platform for the horizontal integration of quality information over the complete supply chain has to face three major challenges:

1. Guarantee reliability of quality information
2. Supervise, that the product and the available quality information fulfil given specifications
3. Customize the quality data exchange to contain all relevant information for each individual order

---

[2] http://www.siemens.com/global/en/home/company/topic-areas/digitalization/mindsphere.html
[3] https://www.ge.com/digital/predix
[4] https://www.ibm.com/watson/

A solution realizing Quality4.0 has to respect all these challenges in an integrated way to enable the evolution of efficient and quality-oriented future manufacturing.

## 3  The Quality4.0 approach

To realize the vision of horizontal integration of quality information through the complete supply chain, it is essential to guarantee quality data plausibility, to allocate products (coils, billets, bars) and data to customers/orders, and finally to compile and exchange the data. Therefore, the solution that will be developed within this project will consider all of these aspects, create methods for plausibility protection as well as for quality data management and compile the developed solutions into the Quality4.0 platform shown in Fig. 1. Using a Service-Oriented-Architecture (SOA) the single components of the Quality4.0 platform can be flexible combined and integrated in the existing IT infrastructure without relying on a single software vendor, product or technology.

Fig. 1 shows the concept from the rolling point of view receiving slabs from a steel supplier and providing coils to a customer. . The same concept can be applied to the steelmaking process for example.To respond to the considered needs of the project, the Quality 4.0 Platform will be composed with 3 service components (as illustrated in the  Fig. 1 in framed bold red):

- Quality Data Generation Service (QGS) component generates the quality data and their plausibility values
- Quality Allocation Service (QAS) component allocates the customer order to a suitable product and selects the relevant associated quality data
- Quality Exchange Service (QXS) component exchanges the selected quality data compiled for each customer order

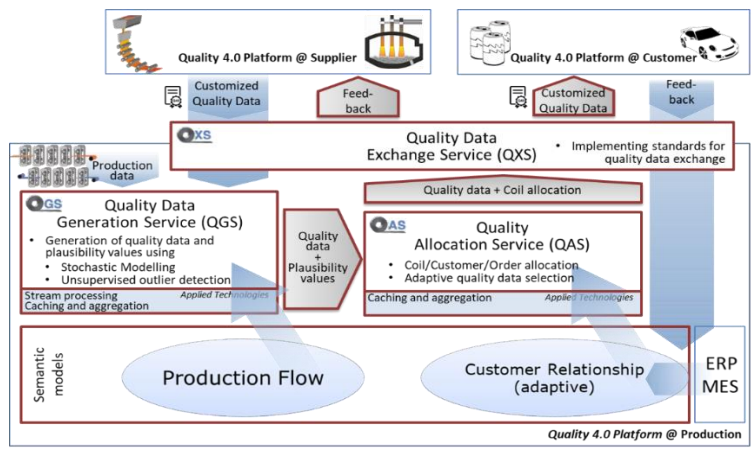

**Fig. 1.** Concept of Quality4.0 architecture (Rolling example)

## 4 Quality Data Generation Service (QGS)

### 4.1 Data Plausibility Protection

In literature Commonly high-quality data is defined from the user's point of view as data that are "fit for use by data consumer" [4]. Reliability of data is seen as a "multidimensional, complex, morphing and goal-oriented concept" [5]. Consequently, the dimensions that have to be considered for data quality evaluation are specific for each data application and there is no common agreement in literature, which dimensions to choose in this definition. In [6] four dimensions, "Accuracy, Completeness, Timeliness, Consistency" have been discovered with major relevance in the context of industrial data quality assessment. This fact is taken into consideration by the papers dealing with the quality of industrial data. Examples from the oil and gas industry are given in [7] and [8], from the chemical industry in [9].

In the field of big data analytics, the consideration of the data quality is just at the beginning, as further challenges have to be met regarding data volume and diversity [10]. In this case increasing demands for the timeliness and the real-time processing of the data is foreseen, where solutions are still lacking.

### 4.2 Plausibility Value Calculation

The main functionality of the Quality4.0-QGS is to estimate quality data from all available data sources and quantify the confidence of this estimation by means of a plausibility value (PV) to final guarantee reliability of quality information provided.

In the daily production aspects like accuracy, completeness, timeliness and consistency of quality related data have to be considered comprehensively for PV calculation. Particularly demanding in this context is the plausibility protection for data coming from automatic surface inspection systems (ASIS), as the quantification of ASIS accuracy is still a subject of ongoing research [11].

As a generic approach for the calculation of PVs the determination of plausibility values can be described as a function

$$PV \coloneqq f(I) \to [0,1] \qquad (1)$$

In the context of this paper a value of 0 always means that the data is not plausible and a value of 1 stands for data which should be fully reliable. Depending on the investigated type of quality data, different input data domains may apply to $I$. Methods covering one dimensional input data ($I = \mathbb{R}$) can be applied directly to 1D quality data as well as indirectly to measurement position or area values originating from 2D quality measurements.

Table 1 shows a list of data plausibility measures used in the Quality4.0 project. These methods are based on simple mathematical formulas and can be combined to a complex plausibility assessment.

**Table 1.** List of data plausibility measures

| | Function | Parameter |
|---|---|---|
| Constant | $f(x) := p_m \in [0,1]$ | $p_m \in [0,1]$ |
| Thresholding | $f(x, t_{min}, t_{max}) := \begin{cases} 0 & \text{for} \quad x < t_{min} \\ 1 & \text{for} \quad t_{min} \leq x \leq t_{max} \\ 0 & \text{for} \quad x > t_{max} \end{cases}$ | $t_{min}, t_{max} \in I$ |
| Fuzzy-based | $f(x, t_0, t_1, t_2, t_3) := \begin{cases} 0 & \text{for} \quad x \leq t_0 \\ \dfrac{x - t_0}{t_1 - t_0} & \text{for} \quad t_0 < x < t_1 \\ 1 & \text{for} \quad t_1 < x \leq t_2 \\ \dfrac{t_3 - x}{t_3 - t_2} & \text{for} \quad t_2 < x \leq t_3 \\ 0 & \text{for} \quad x > t_3 \end{cases}$ | $t_0, t_1, t_2, t_3 \in I$ |
| Variation-based | $f(x) := \begin{cases} 0 & \text{for} \quad \max(W_x^n) - \min(W_x^n) = 0 \\ 1 & \text{otherwise} \end{cases}$ | $n \in \mathbb{N}$ |
| Data driven | $f(x) := PV(x)$ | $PV: I \to [0,1]$ |

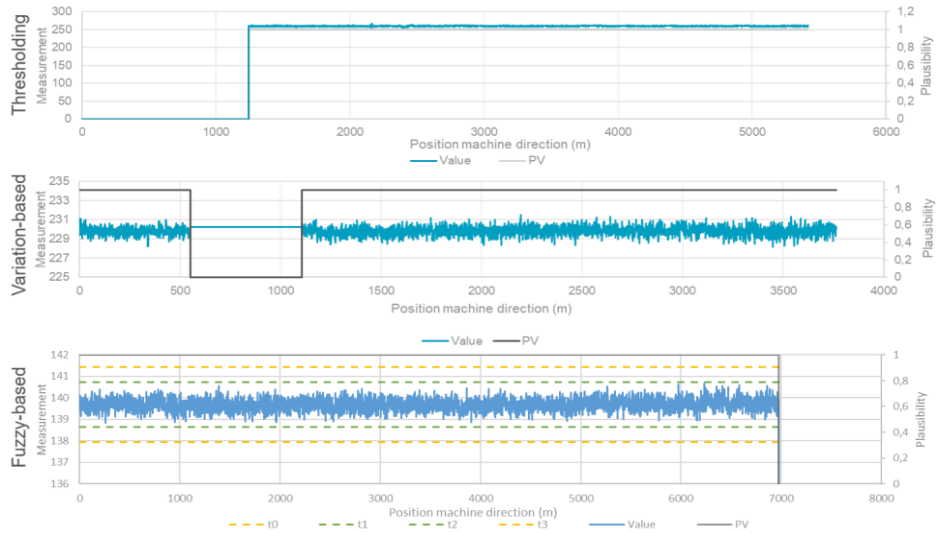

**Fig. 2.** Examples of position-based plausibility value calculation at a rolling mill

### 4.3 Outlier detection

Quality related data collected throughout the production process are subjected to malfunctions in the sensors performing the measurement or other potential sources of errors in the measurements that lead to the creation of outliers. An informal definition of the concept of outlier involves its deviation from normality. There are multiple more formal definitions of outliers in literature that depend on the context [12] and can be grouped into 5 categories: distribution based, depth based, distance based, clustering based and density based. Each of these group focuses on a particular aspect of the deviation from normality.

In the context of the Quality4.0 project an efficient and reliable detection of outlier quality measures plays a fundamental role as it avoids the sharing of misleading information about the sold products with customers that would degrade the customer-producer trust relationship. For this reason a module for the outlier-level calculation was embedded into the data plausibility component of the framework: this module is able to associate to each quality measure sample a outlier-risk value in the range [0;1] in a completely automatic manner and thus realizes a data-driven PV calculation $f(x) = PV(x)$ as described in the previous section.

Due to the wide typology and variety of outliers there is not a commonly recognized approach for their detection that is efficient and reliable in any context. This fact is related to the above-mentioned different categories of definition that can be provided for the concept of outlier itself. Within the project the FUCOD [13] algorithm for outliers detection was used. This method combines 4 existing approaches to outliers detection with the aim of exploiting their strong points and avoiding their weaknesses according to processed data by means of a Fuzzy Inference System (FIS) that manages dynamically the contribution of each single method. FUCOD is designed to work with multidimensional data which means that the outlier-level will not be calculated only by considering the characteristics of the individual variables that form the quality data but also the interactions among them within the problem domain. These characteristics make the FUCOD approach particularly suitable to cope with industrial dataset as put into evidence by the numerous tasks, even within the steelmaking industry, where the method was successfully applied [14][15].

As shown in Fig. 3 FUCOD exploits 4 different methods belonging to different outlier detection algorithms families, namely:

- The Grubb's test, belonging to distribution-based methods, that takes into account the data deviation from a normal distribution
- A distance-based indicator that considers the distribution of the distances of data from their neighbors
- A clustering indicator that assesses the membership level of each data to its data-cluster within the dataset
- The Local Outlier Factor (LOF), a density-based approach that takes into account the density of data in the dataset domain

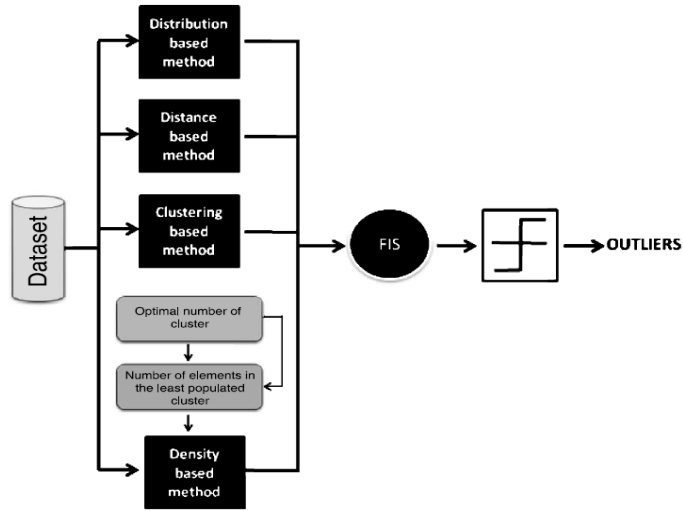

**Fig. 3.** Flowchart depicting the FUCOD algorithm for outliers detection

The output of single methods is suitably mixed by a FIS formed by a set of 6 rules and that returns, as final outcome, the outlier-level of each single observation in the dataset. In Quality4.0 FUCOD is applied to the detection of outliers in data related to sensors measurements on manufactured coils. More in detail two different data sources were used in association to each coil: the first one consists of a set of 3 mono-dimensional measures (called P1, P2 and P3) concerning quality properties throughout the length of the coil, the second one (called P4) is a 2-dimensional map that covers the entire product surface. The FUCOD outliers detection was employed on a dataset of observation throughout the length of the coil where each data sample consists of P1, P2, P3 and the average of the P4 value along the coil width. The so-constituted 4-dimensional dataset is fed to FUCOD that automatically calculated the overall outlier level of each data sample corresponding to a set of measures over the coil length.

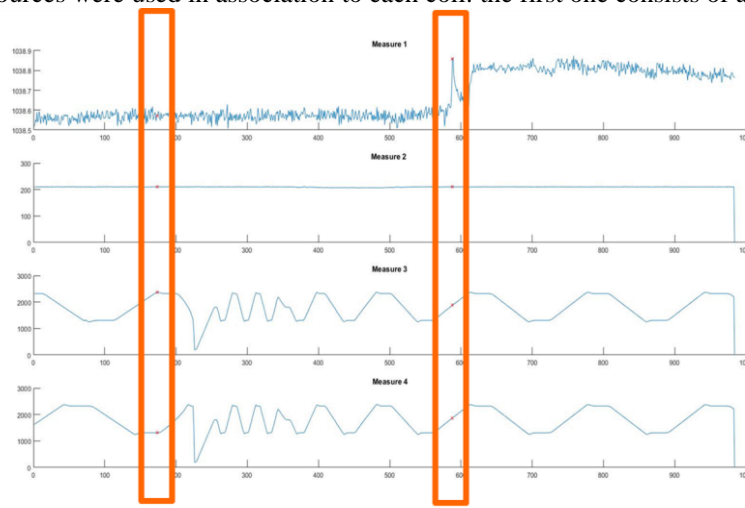

**Fig. 4.** Outcome of the FUCOD method on data belonging to a sample coil. Data samples with higher risk of being outliers are highlighted.

From a qualitative point of view the method achieves satisfactory results as it is able to point out data samples that seems to deviate from standard multi-dimensional distribution of observations in the dataset as depicted in Fig. 4. In general, on the tested coils, FUCOD detects a number of outlier samples that varies in the 0.01%-0.1% with respect to the number of observations within a dataset. The average computational time for each coil is 25 seconds that makes it acceptable to the integration in the Quality4.0 framework.

## 5    Quality Allocation Service (QAS)

The Quality4.0-QAS realizes adaptive supervision of product quality by combination of quality data and knowledge about customer relations. As shown in Fig. 1 the Quality4.0-QGS provides estimated quality data together with a dedicated plausibility value reflecting the confidence of the quality data. The Quality4.0-QAS again receives this information and combines it with knowledge about the intended customer resp. order. Based on this information the Quality4.0-QAS has to decide autonomously if the product can be allocated to this customer resp. order and selects the relevant quality data to exchange. Furthermore, in case of quality deficiencies the information provided by the Quality4.0-QGS will be compiled by the Quality4.0-QAS to create valuable feedback information for the supplier and send it to the QXS.

However, a system that intends to exchange relevant quality information between supplier and customer has to understand the meaning of "relevance". Thus, all information required for the determination of quality information relevance was semantically modelled based on available customer and order data. This customer relationship model implies to reflect the mutual trust between supplier and consumer and comprises the information as well under which conditions a coil fits to a certain order as which kind of quality information the Quality4.0-QXS should provide to the specific customer. The level of customer intimacy was included in the model to be able to define the type, amount and plausibility of quality information that should be exchanged considering the interrelationship between producer and customer.

## 6    Quality Exchange Service (QXS)

Finally, based on the result provided by the Quality4.0-QAS, the selected quality data will be compiled for each order individually and the data will be exchanged using a standard communication protocol. Therefore, the QXS is the only service that is ac-cessible over plant boundaries and manages the quality data exchange between different instances of the Quality4.0 platform. Thus, by exchanging quality and feedback information between supplier and customer, it realizes a fully bi-directional customer-oriented quality data exchange and establishes a common focus on product quality by horizontal integration. To determine a suitable IT standard for quality data exchange between the customers and the suppliers the following list of solutions has been defined and analyzed: the QDX [5], the STEP[6], and the Quality Tracking System[7]. As it turned out that no suitable free standard for quality data exchange exists, a Quality4.0 specific IT standard will be defined and implemented in the Quality4.0 framework.

## 7 Implementation

### 7.1 The FADI framework

FADI is a customizable end-to-end big data platform enabling the deployment and the integration of open source tools in a portable and scalable way. It is a multi-tenant and multi-actors (i.e. business analyst, data scientist/engineer, IT admin, etc.) platform. The main features of FADI are five-fold (see Fig. 5): (1) collecting batch and stream data coming from various data sources, (2) storing data in different types of data stores, (3) processing data using ML and Artificial Intelligence (AI) techniques, (4) visualizing and analyzing data in a user Web interface, and (5) generate and publishing reports.

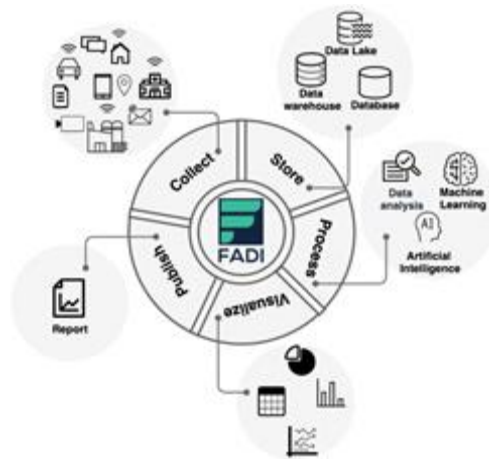

**Fig. 5.** An overview of FADI's features

---

[5] VDA-QDX: https://portal3.gefeg.com/vdaqdx/page/home
[6] http://www.steptools.com/stds/step/
[7] http://www.eurofer.eu/Issues%26Positions/Quality%20Tracking/Quality%20Tracking.fhtml

### 7.2 A sample implementation of FADI

Some of the innovative aspects of FADI is the customizability, modularity and extendibility. Indeed, it provides a flexible software architecture that enables to easily integrate various open source tools dedicated to the management of Big Data and third-party tools (i.e. QGS, QAS, QXS, and Modoboa). Indeed, we present the implementation of FADI that allows to set-up the Quality4.0 platform. In the following, we introduce how the FADI platform has been customized in order to put in place the Quality4.0 and support the requirements of the project (see Fig. 6):

- For the **internal administration services**, we propose to use the following tools: The ELK[8] tools suite is used to manage the logs. ELK stands for the three open source tools: ElasticSearch to search information in the collected logs, Logstash to transport and process logs by collaborating with the Elasticsearch tool, and Kibana to visualize the logs and the results of its analysis. Then, Zabbix[9] is used to set-up the monitoring in the FADI infrastructure and OpenLDAP[10] is used to provide Identity and Access Management of the different users of the platform. Finally, we propose to configure the server mail Modoboa[11] in order to enable the exchange of the Quality4.0 certificate and the feedback between the customer and the producer via the QXS.
- We propose to use the Apache Nifi[12] tool to set up the **data integration** layer. It automates the data flow between systems: both with external systems (e.g. sensors data sources) and between internal systems (e.g. the various analysis and storage components). Apache Nifi is connected to the QGS in order to compute the PVs, to detect the outliers and to store raw and computed data.
- For the **data analysis** layer, we propose to use and integrate in the FADI platform the QGS and the QAS components. The QGS computes the different metrics. Whereas the QAS evaluates the quality of a given product and prepares the Quality4.0 certificate.
- For the **data storage** layer, we propose to use the PostgreSQL[13] which is an object-relational database management system. It has been chosen because it contains advanced database features (e.g. aggregation, parallel queries, replication system, etc. This database will be used in order to store all the data (e.g. the production flow, customer relationship, ERP/MES, etc.) and the semantic models.
- For the **serving layer**, we propose to use Adminer[14] to manage the PostgreSQL database in a web interface. Grafana[15] is used to explore and visualize data. Then, the QAS interface is available to control the data quality allocation. Finally, the

---

[8] https://www.elastic.co/what-is/elk-stack

[9] https://www.zabbix.com/

[10] https://www.openldap.org

[11] https://modoboa.readthedocs.io/en/latest/

[12] https://nifi.apache.org/

[13] https://www.postgresql.org/

[14] https://www.adminer.org/

[15] https://grafana.com/

QXS interface is used to send and receive the Quality4.0 certificate and the customer feedbacks.

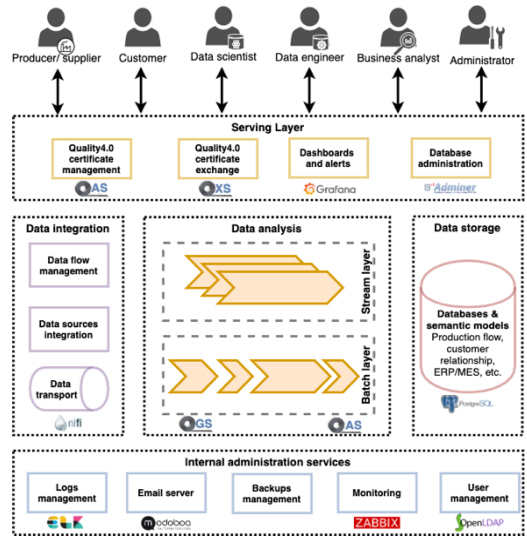

**Fig. 6.** Quality4.0 platform implemented using the FADI platform

### 7.3 The CI/CD pipeline of FADI

To ensure speed and quality of the system's lifecycle, the Quality 4.0 platform relies on the DevOps approach that combines development and operations processes: plan, code, build, test, release, deploy, operate, monitor. Those processes are automated through Continuous Integration and Delivery (CI/CD) pipelines to facilitate frequent iterations on integration between the platform components (e.g. QAS, QGS, QXS, data sources, data sinks, etc.) and the deployment of the solution to various environments (e.g. testbeds, servers in factories, etc.). Fig. 7 illustrates the services and technologies that support the DevOps processes of the platform:

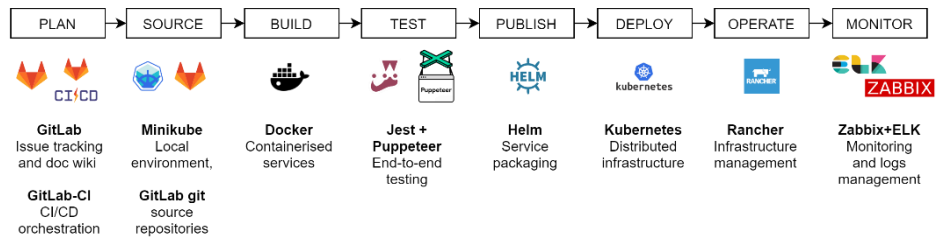

**Fig. 7.** Quality 4.0 DevOps CI/CD processes

## 8     Conclusion

Only if the European Steel Industry succeeds to win customer-trust and solidifies client intimacy, it will achieve a durable competitive advantage and thus reduce pressure from world-wide imports. Furthermore, it is of strategic importance for the European steel industry to proactively promote such a common platform instead of reacting on specific customer demands. In the RFCS project Quality4.0 presented in this paper an adaptive platform tailored to the demands of the steel industry will be developed that realises the horizontal integration of quality information across the entire supply chain. In detail this new platform will provide quality information of high reliability, improved decisions on the reached product quality and automatic data exchange adapted to existing customer and order information.

## References

1. Bonnaud, E., De Thieulloy, G., Lecat, A., Saint-Aubyn, J., Charpentier, A.: THINK ACT - Weathering the steel crisis, Paris: Roland Berger (2016).
2. acatech – Deutsche Akademie der Technikwissenschaften e.V., Umsetzungsempfehlungen für das Zukunftsprojekt Industrie 4.0, Frankfurt Main (2013).
3. IW Consult GmbH: Potentiale des digitalen Wertschöpfungsnetzes Stahl, WV Stahl (2017)
4. Wang, R.Y., Strong, D. M.: Beyond Accuracy: What Data Quality Means to Data Consumers, Journal of Management Information Systems, Vol. 12, No. 4 (1996)
5. Dasu, T. Johnson, T.: Exploratory Data Mining and Data Cleaning, John Wiley (2003)
6. Gertz, M., Tamer Ozsu, M., Saake, G., Sattler, K.: Data Quality on the Web, Report on the Dagstuhl Seminar (2004)
7. Radhay,R. Schlumberger: Facilitating data quality improvement in the oil and gas sector, SPE Asia Pacific Oil and Gas Conference and Exhibition, Perth, Australia (2008)
8. Hubauer,T., Lamparter, S., Roshchin, M., Solomakhina, N., Watson, S.: Analysis of data quality issues in real-world industrial data, Conference of the Prognostics and Health Management Society, New Orleans, USA (2013)
9. Jiang, T.W., Stuart, P.R., Chen, B. and Jasim,K.: Strategy for Improving Data Quality for a Kraft Pulp Mill Recausticizing Plant, Proc. FOCAPO Conf., Coral Springs Florida (2003)
10. Cai, L and Zhu, Y: The Challenges of Data Quality and Data Quality Assessment in the Big Data Era. Data Science Journal (2015)
11. J. Brandenburger, H. Krambeer, C. Schirm, K. Jonker, W. Mißmahl, M. Nörtersheuser, A. Kogler, A. Ebner: Towards measuring surface quality by means of automatic surface inspection systems, 4th European Steel Technology and Application Days, Düsseldorf (2019)
12. V. Barnett, T. Lewis: *Outliers in Statistical Data*, Wiley Series in Probability and Mathematical Statistics, John Wiley & Sons; Chichester, 1994.
13. S. Cateni, V. Colla, M. Vannucci, *A fuzzy logic-based method for outliers detection*, In Artificial Intelligence and Applications, February 2007, pp. 605-610
14. Dimatteo, A., Vannucci, M., Colla, V., *Prediction of hot deformation resistance during processing of microalloyed steels in plate rolling process*, In The International Journal of Advanced Manufacturing Technology, *66*(9-12), 1511-1521, (2013).
15. V. Colla, G. Bioli, M. Vannucci, Model parameters optimisation for an industrial application: A comparison between traditional approaches and genetic algorithms, Proceedings - EMS 2008, European Modelling Symposium, art. no. 4625243, pp. 34-39, (2008).